\documentclass[preprint,12pt]{elsarticle}

\usepackage{amssymb}
\usepackage{amsmath}
\usepackage{booktabs}
\usepackage{tikz}
\usepackage{float}
\usepackage{multirow}
\usepackage{color}
\usepackage{xcolor}
\usepackage{hyperref}
\usepackage{array}
\usepackage{rotating} % sidewaysfigure / sidewaystable 제공

\AtBeginEnvironment{tabular}{\scriptsize}
\AtBeginEnvironment{tabularx}{\scriptsize}
\AtBeginEnvironment{longtable}{\scriptsize}

\journal{NeuroComputing}

\begin{document}

\begin{frontmatter}

\title{Disentangled Concept Representation for Text-to-image Person Re-identification}

% \author{Anonymous authors}

% 저자: (1) 제1저자 (2) 제2저자 (3) 제3저자 (4) 교신저자
\author[inst1]{Giyeol Kim}            % 제1저자
\ead{giyeolkim@cau.ac.kr}

\author[inst2]{Chanho Eom\corref{cor1}}% 교신저자
\ead{cheom@cau.ac.kr}

\affiliation[inst1]{%
  organization={Department of Imaging Science, Graduate School of Advanced Imaging Science, Multimedia \& Film, Chung-Ang University},
  addressline={},
  city={Seoul},
  postcode={06974},
  state={},
  country={South Korea}
}

\affiliation[inst2]{%
  organization={Department of Metaverse Convergence, Graduate School of Advanced Imaging Science, Multimedia \& Film, Chung-Ang University},
  addressline={},
  city={Seoul},
  postcode={06974},
  state={},
  country={South Korea}
}

\cortext[cor1]{Corresponding author \\ 
Full postal address: Graduate School of Advanced Imaging Science, Multimedia \& Film, Chung-Ang University, Seoul, 06974, Korea.}

% use optional labels to link authors explicitly to addresses:
% \author[label1,label2]{}
% \affiliation[label1]{organization={},
%             addressline={},
%             city={},
%             postcode={},
%             state={},
%             country={}}
%
% \affiliation[label2]{organization={},
%             addressline={},
%             city={},
%             postcode={},
%             state={},
%             country={}}

%% Abstract

\begin{abstract}
Text-to-image person re-identification (TIReID) aims to retrieve person images from a large gallery given free-form textual descriptions. TIReID is challenging due to the substantial modality gap between visual appearances and textual expressions, as well as the need to model fine-grained correspondences that distinguish individuals with similar attributes such as clothing color, texture, or outfit style. To address these issues, we propose DiCo (Disentangled Concept Representation), a novel framework that achieves hierarchical and disentangled cross-modal alignment. DiCo introduces a shared slot-based representation, where each slot acts as a part-level anchor across modalities and is further decomposed into multiple concept blocks. This design enables the disentanglement of complementary attributes (\textit{e.g.}, color, texture, shape) while maintaining consistent part-level correspondence between image and text. Extensive experiments on CUHK-PEDES, ICFG-PEDES, and RSTPReid demonstrate that our framework achieves competitive performance with state-of-the-art methods, while also enhancing interpretability through explicit slot- and block-level representations for more fine-grained retrieval results.
\end{abstract}

% %Graphical abstract
% \begin{graphicalabstract}
%     \includegraphics[width=\textwidth]{grahpical_abstract.pdf}
% \end{graphicalabstract}

% %Research highlights
% \begin{highlights}
% \item Proposing DiCo to achieve hierarchical cross-modal alignment for TIReID
% \item Designing slot-concept attention to disentangle features into interpretable concepts
% \item Demonstrating competitive results on three benchmarks through extensive experiments
% \end{highlights}

% Keywords

\end{frontmatter}

%% Use \section commands to start a section
\section{Introduction} \label{sec:intro}

Text-to-image person re-identification (TIReID) aims to retrieve a specific individual from a large-scale image gallery given only a natural language description~\cite{chen2018improving, li2017identity, li2017person, chen2021cross, sarafianos2019adversarial, zhang2018deep, chen2022tipcb, ding2021semantically, gao2021contextual, niu2020improving, park2024plot, suo2022simple, JI2018205, WANG2021388, kim2025leveraging}. Unlike traditional image-based ReID~\cite{he2021transreid, luo2019bag, ZHANG2025108, ZHANG202466, WANG2023237, CHEN202290, eom2025cerberus, lee2025domain, eom2019learningdisentangledrepresentationrobust, ZHANG2023453, SARKER2025129257, spc, ding2025decoupling, ding2025person, ding2024clothes}, TIReID enables queries to be expressed directly in natural language, offering a more flexible and practical retrieval paradigm. Despite its practical advantages, TIReID remains highly challenging due to two key issues: 1) the substantial modality gap between visual and textual representations, which arises from their inherently heterogeneous forms of expression. Images consist of dense and continuous visual signals that capture appearance variations such as pose, viewpoint, and illumination, whereas textual descriptions rely on discrete linguistic tokens to convey high-level semantic attributes. This discrepancy makes it difficult to directly align the two modalities, highlighting the necessity of learning a unified embedding space for reliable cross-modal matching. 2) Another challenge lies in modeling fine-grained correspondences between textual attributes and subtle visual cues in images. Since part-level annotations are rarely available in real-world datasets, it is essential to develop models that can automatically discover and align discriminative local features without relying on explicit supervision.

\begin{figure*}[t!]
  \centering
  \includegraphics[width=\linewidth]{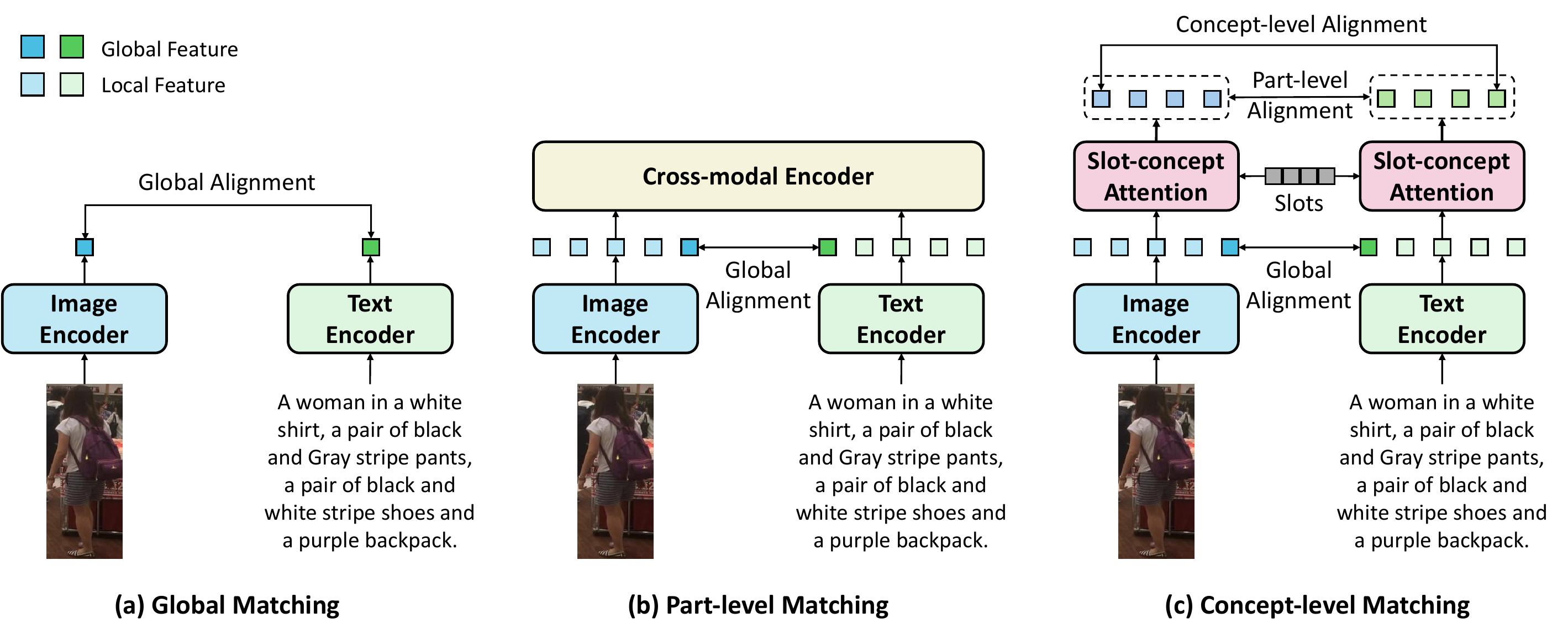}
  \caption{
  Illustration of alignment strategies for TIReID. 
  \textbf{(a) Global matching:} aligns image and text only at the global level, missing local details. 
  \textbf{(b) Part-level matching:} introduces region–word alignment but still entangles multiple attributes within each part. 
  \textbf{(c) Concept-level matching (Ours):} disentangles slots into concept-specific blocks, enabling hierarchical alignment from global identity cues to fine-grained attributes such as color, texture, and shape. 
  }
  \label{fig:teaser}
  \vspace{-1em}
\end{figure*}

To address these challenges, early methods~\cite{chen2018improving, li2017identity, li2017person, chen2021cross, sarafianos2019adversarial, zhang2018deep} typically adopt dual-encoder frameworks, where images and texts are projected into a shared embedding space using global representations and metric learning objectives (Fig.~\ref{fig:teaser} (a)). However, such global alignment often fails to capture discriminative local details. To overcome this limitation, part-based and region-level methods~\cite{chen2022tipcb, ding2021semantically, gao2021contextual, niu2020improving, park2024plot, suo2022simple, WANG2021388} decompose person images into body parts and align them with word-level textual tokens to establish localized correspondences (Fig.~\ref{fig:teaser} (b)). More recently, vision-langauge model (VLM)~\cite{CLIP} is leveraged to enhance textual understanding and contextual reasoning, enabling richer semantic representations that facilitate more accurate retrieval~\cite{jiang2023cross, yan2023clip, park2024plot}. Despite these advances, existing methods still struggle to fundamentally bridge the heterogeneous nature of visual and textual modalities. To mitigate this problem, prior methods~\cite{chen2018improving, li2017identity, li2017person, chen2021cross, sarafianos2019adversarial, zhang2018deep, chen2022tipcb, ding2021semantically, gao2021contextual, niu2020improving, jiang2023cross, yan2023clip, park2024plot} attempt to narrow the modality gap by either aligning dual-encoder features in a shared embedding space through metric objectives or introducing cross-attention mechanisms on top of dual-encoder representations. Although these strategies allow cross-modal interactions, they often result in shallow feature fusion rather than deep semantic alignment. Moreover, alignment in prior methods~\cite{chen2022tipcb, ding2021semantically, gao2021contextual, niu2020improving, park2024plot, suo2022simple} is typically restricted to global identity features or coarse part-level correspondences, which cannot explicitly disentangle fine-grained semantic concepts such as color, texture, or shape. Consequently, subtle yet discriminative cues that are critical for distinguishing visually similar individuals may be frequently overlooked.

In this paper, we propose DiCo (Disentangled Concept Representation), a novel framework that models hierarchical and interpretable cross-modal alignments through structured slot decomposition and concept disentanglement (Fig.~\ref{fig:teaser} (c)). DiCo introduces a unified set of learnable slots that act as modality-shared anchors across body regions, and further factorizes each slot into multiple concept blocks capturing semantically coherent attributes such as color, texture, and shape. This slot-concept structure enables DiCo to bridge the inherent modality gap not by shallow fusion, but through iterative and attentive interactions that gradually refine semantic consistency across modalities. Moreover, by organizing the representation space hierarchically from global identity cues to concept-specific subspaces, DiCo can automatically localize and align subtle visual attributes with their textual counterparts even without relying on explicit part-level annotations. This design allows our model to reason over both coarse and fine-grained semantics, significantly enhancing its ability to distinguish individuals with subtle attribute variations. By supervising alignment at global, part, and concept levels through multi-level contrastive objectives and reconstruction constraints, DiCo learns robust and semantically grounded embeddings. Extensive experiments on three public benchmarks validate the effectiveness of DiCo, achieving competitive performance against state-of-the-art methods.

The main contributions of this work are as follows:

\begin{itemize}
\item We propose DiCo, a novel framework that introduces a unified slot-based decomposition with concept-level factorization, enabling hierarchical alignment across global identity cues, body regions, and fine-grained semantic concepts such as color, texture, and shape.
\item We design a slot-concept attention mechanism that jointly learns modality-shared part anchors and disentangled concept blocks, allowing for interpretable and robust cross-modal matching without requiring explicit part-level annotations.
\item We demonstrate the effectiveness of DiCo through extensive experiments on CUHK-PEDES, ICFG-PEDES, and RSTPReid, where it achieves competitive performance against state-of-the-art methods.
\end{itemize}

\section{Related Works} \label{sec_rw:2.1}

\subsection{Text-to-image Person Re-identification} \label{subsec:Text-to-image Person Retrieval}
Text-to-image person re-identification (TIReID) aims to retrieve the correct person image from a large gallery based on a free-form textual description. Early methods~\cite{chen2018improving, li2017identity, li2017person, chen2021cross, sarafianos2019adversarial, zhang2018deep} typically adopt dual-encoder architectures, where images and texts are embedded into a shared space using global alignment objectives. While such methods establish a foundation for cross-modal retrieval, they struggle to capture fine-grained attributes essential for distinguishing visually similar individuals. To address this issue, subsequent methods~\cite{chen2022tipcb, ding2021semantically, gao2021contextual, niu2020improving, park2024plot, suo2022simple, WANG2021388} introduce region- and part-based models that decompose person images into body regions and align them with word- or phrase-level tokens, enabling more localized correspondences. Other methods~\cite{aggarwal2020text, jing2020pose, wang2020vitaa} leverage auxiliary cues such as pose estimation, attribute prediction, or human parsing to provide additional structural priors. More recently, several methods~\cite{jiang2023cross, yan2023clip, park2024plot} adopt vision–language models~\cite{CLIP} to enhance textual understanding and contextual reasoning, enabling the capture of richer semantic attributes in TIReID.

Despite these advances, existing methods still face fundamental limitations in modeling fine-grained cross-modal correspondences. While region- and part-based methods~\cite{chen2022tipcb, ding2021semantically, gao2021contextual, niu2020improving, park2024plot, suo2022simple, WANG2021388} have enabled more localized alignment, the extracted part features are typically entangled representations that fail to separate key semantic attributes like color, texture, and shape within each region. Such entangled representations hinder precise alignment, as they obscure subtle yet discriminative cues necessary for identifying visually similar individuals. These challenges highlight the need for a more structured framework that can capture hierarchical and semantically disentangled correspondences across modalities.

\subsection{Slot Attention} \label{subsec:2.2}
Slot Attention~\cite{slot_attention} is proposed for unsupervised object-centric learning, aiming to decompose visual scenes into slot-based latent representations through an iterative attention mechanism. Each slot typically corresponds to an object or a part, yielding structured and interpretable representations without requiring explicit supervision. Building on this idea, subsequent methods extend Slot Attention to capture temporal dynamics in videos~\cite{kipf2021conditional, fan2023unsupervised, elsayed2022savi++, singh2022simple}, or integrate top-down signals to improve stability and semantic alignment~\cite{kim2024bootstrapping}. More recently, Slot Attention has also been explored in cross-modal contexts~\cite{kim2023improving, didolkar2025ctrl, kim2023shatter} and Neural Concept Binders~\cite{stammer2024neuralconceptbinder} and Systematic Binders~\cite{singh2023neuralsystematicbinder} leverage slot-based architectures to disentangle input representations into structured concept factors, enabling more compositional and interpretable reasoning. While slots provide region-level representations, the semantic concepts within each region remain entangled, which limits their ability to capture fine-grained correspondences. In contrast, our DiCo extends the slot-based paradigm by introducing modality-shared slots that serve as common anchors across vision and language, and further factorizing each slot into disentangled concept representations. This design enables attributes such as color, texture, and shape to be aligned independently while maintaining consistent cross-modal grounding.

\subsection{Multi-granularity Representation Learning} \label{subsec:Multi-granularity}
Multi-granularity representation learning has been widely explored as an effective strategy for capturing complementary information at different semantic levels, ranging from global context to fine-grained local details. In the person re-identification and person search domains, multi-granularity learning has proven particularly effective for handling subtle appearance differences among visually similar individuals. By jointly modeling representations at multiple granularities, such methods improve robustness to variations in scale, pose, and appearance while enhancing the discriminative power of learned features. Recent works have investigated hierarchical architectures that explicitly integrate multi-level representations through multi-scale feature extraction~\cite{gao2019res2net, gong2021lag, zhang2025mixed, yang2021learning, osnet}, pyramid structures~\cite{yan2020learning}, or granularity-aware attention mechanisms~\cite{chen2019abd}. Recent studies~\cite{feng2021homogeneous, feng2025homogeneous} further introduce structured and multi-granularity representations to align heterogeneous features across modalities, demonstrating that granularity-aware modeling plays a crucial role in improving cross-domain robustness and fine-grained alignment. Despite these advances, most existing methods primarily focus on coarse granularity decomposition, where fine-grained semantic attributes within each region remain entangled. This limitation motivates the need for more structured multi-granularity frameworks that can explicitly disentangle semantic concepts while preserving consistent correspondence across different representational levels.

\begin{figure*}[t!]
  \centering
  \includegraphics[width=\linewidth]{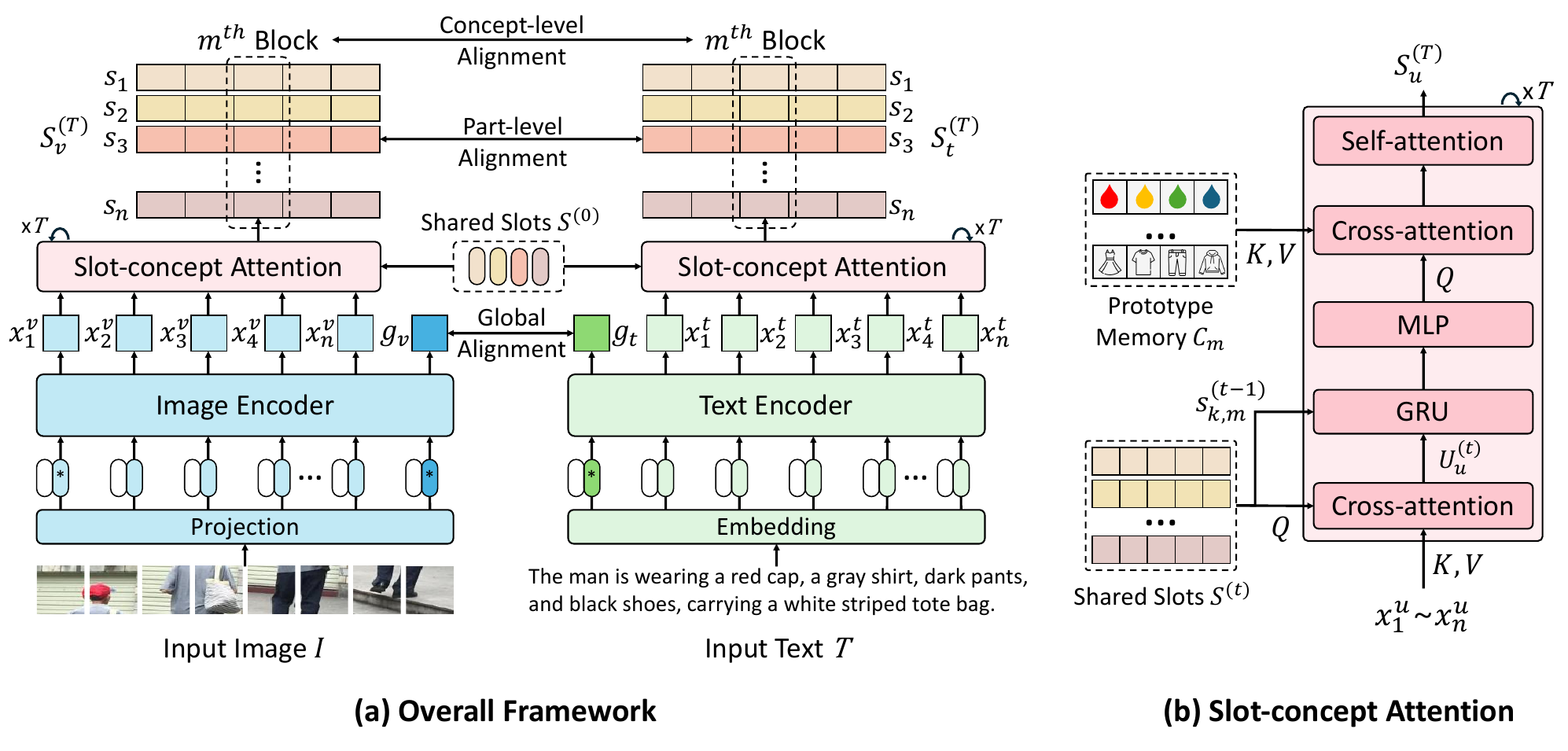}
  \caption{
  \textbf{(a) Overall framework:} Given an input image and text, visual and textual features are extracted by respective encoders, followed by global alignment and refinement through shared slots. The refined slot representations are aligned at both part- and concept-levels to capture fine-grained cross-modal correspondences. 
  \textbf{(b) Slot-concept Attention:} disentangles slot representations into interpretable concept blocks while ensuring semantic consistency across modalities.
  }
  \label{fig:main}
  \vspace{-1em}
\end{figure*}

\section{Proposed Method} \label{sec:proposed_method}
We propose DiCo (Disentangled Concept Representation), which learns hierarchical and semantically disentangled cross-modal correspondences for text-to-image person re-identification. The overall architecture is introduced in Sec.\ref{subsec:3.1}, the slot-concept disentanglement mechanism is detailed in Sec.\ref{subsec:3.2}, and the training and inference strategies are discussed in Sec.~\ref{subsec:3.3}.

\subsection{Overall Framework} \label{subsec:3.1}
As illustrated in Fig.~\ref{fig:main}(a), the proposed framework processes an input image--text pair from bottom to top and gradually builds three levels of representation: global, part-level, and concept-level.

\paragraph{Encoders}
Given a person image $I$ and a free-form textual description $T$, the goal of TIReID is to retrieve person images that match the given text by learning discriminative and semantically aligned cross-modal representations.
For the visual branch, a visual backbone first splits $I$ into patches and projects them into a sequence of patch tokens
$\mathbf{X}_v = \{ x_n^v \}_{n=1}^{N} \in \mathbb{R}^{N \times d}$,
together with a global representation $g_v \in \mathbb{R}^{d}$ (\textit{e.g.}, the \texttt{[CLS]} token in a transformer).
Here, $N$ denotes the number of image patches and $d$ is the feature embedding dimension.
Similarly, the textual branch encodes the description $T$ into word or subword tokens
$\mathbf{X}_t = \{ x_\ell^t \}_{\ell=1}^{L} \in \mathbb{R}^{L \times d}$,
and a global representation $g_t \in \mathbb{R}^{d}$, where $L$ denotes the number of text tokens.

\paragraph{Shared slots and initialization}
To achieve fine-grained cross-modal alignment, we introduce a shared set of $K$ learnable slots
$\mathbf{S}^{(0)} = \{ s_k^{(0)} \}_{k=1}^{K}$, which are used by both modalities.
Each slot is initialized as $s_k^{(0)} \in \mathbb{R}^{M \times d_c}$ and is further factorized into $M$ concept blocks such that
$s_k = [\,s_{k,1}; \dots; s_{k,M}\,]$ with $s_{k,m} \in \mathbb{R}^{d_c}$.
Sharing the same $k$-th slot across the visual and textual branches encourages it to capture a consistent semantic body part (\textit{e.g}., upper body, legs, or shoes) in the image and the corresponding descriptive phrase in the text.
Meanwhile, the concept blocks within each slot provide a disentangled representation of complementary attributes such as color, texture, or shape.

\paragraph{Slot-concept refinement}
Starting from $\mathbf{S}^{(0)}$, the modality-specific tokens $\mathbf{X}_v$ and $\mathbf{X}_t$ are fed into the slot-concept attention modules (one for each modality in Fig.~2(a)).
At each refinement step, the slots act as queries and pull relevant information from the token sequences via cross-attention, aggregate the attended features, and update their representations.
This iterative interaction is repeated $T$ times so that each slot progressively specializes to a particular body region while its internal blocks specialize to different semantic factors within that region.
After $T$ refinement steps, we obtain the final slot embeddings for the visual and textual branches:
\begin{equation}
\mathbf{S}_v^{(T)} = \{ s_k^{v,(T)} \}_{k=1}^{K}, 
\qquad
\mathbf{S}_t^{(T)} = \{ s_k^{t,(T)} \}_{k=1}^{K},
\end{equation}
where $s_k^{v,(T)}$ and $s_k^{t,(T)}$ denote the $k$-th refined slot, each still decomposed into $M$ concept blocks. Further details are described in Sec.~\ref{subsec:3.2}.

\paragraph{Multi-level alignment}
The model outputs both global embeddings $\big(g_v, g_t\big)$ and refined slot embeddings $\big(\mathbf{S}_v^{(T)}, \mathbf{S}_t^{(T)}\big)$.
During training, these representations are supervised with multi-level objectives:
(i) global alignment between $g_v$ and $g_t$ to capture identity-level semantics,
(ii) part-level alignment across slots $\mathbf{S}_v^{(T)}$ and $\mathbf{S}_t^{(T)}$ to enforce consistent body-region correspondence, and
(iii) concept-level alignment between the blocks inside each slot to match fine-grained attributes such as colors or clothing types.
During inference, the final similarity between a text query and a gallery image is computed as a weighted combination of
global similarities (identity cues),
slot-level similarities (region-level alignment),
and block-level similarities (concept-level matching),
corresponding respectively to the three levels depicted in Fig.~2(a).

\subsection{Slot-concept Refinement} \label{subsec:3.2}

Fig.~2(b) shows how the slot-concept attention module gradually refines the shared slots by interacting with image and text tokens. The key idea is that each slot learns to focus on a
particular region of the person (\textit{e.g.}, head, upper body, legs), while each
concept block within a slot captures complementary details such as color,
texture, or clothing type. By gradually interacting with visual and textual
tokens, the slots specialize into consistent and interpretable concepts across
both modalities.

At refinement step $t$, input tokens from each modality are assigned to slots through a cross-attention mechanism. Specifically, for modality $u \in \{v, t\}$, we obtain queries from the slots $q_u(\mathbf{S}^{(t-1)}) \in \mathbb{R}^{K \times d_h}$ and keys/values from the input tokens $k_u(\mathbf{X}_u), v_u(\mathbf{X}_u) \in \mathbb{R}^{|\mathbf{X}_u| \times d_h}$, where $q_u$, $k_u$, and $v_u$ denote modality-specific learnable linear projections into the shared slot dimension $d_h$. The attention score matrix is then computed as
\begin{equation}
\mathbf{M}_u^{(t)} =
\frac{k_u(\mathbf{X}_u)\, q_u(\mathbf{S}^{(t-1)})^\top}{\sqrt{d_h}}
\in \mathbb{R}^{|\mathbf{X}_u|\times K},
\end{equation}
which represents the similarity between each input token and each slot.
We then apply softmax over slots, meaning each token chooses which
slot it belongs to:
\begin{equation}
\mathbf{A}_u^{(t)} = \mathrm{softmax}_k(\mathbf{M}_u^{(t)}).
\end{equation}
This design forces slots to compete for tokens. As a result, each slot
becomes responsible for different parts of the image or different phrases in
the text, rather than all slots redundantly attending to everything. A
normalization step is additionally applied to avoid one slot dominating all
tokens.

Using the attention weights, we aggregate relevant information and obtain 
per-slot feature summaries:
\begin{equation}
\mathbf{U}_u^{(t)} =
\left(\tilde{\mathbf{A}}_u^{(t)}\right)^\top v_u(\mathbf{X}_u).
\end{equation}
Each row of $\mathbf{U}_u^{(t)}$ corresponds to one slot, and is further divided
into $M$ concept blocks. These blocks are refined independently using a
lightweight GRU and MLP:
\begin{equation}
\hat{s}_{k,m}^{(t)} =
\mathrm{GRU}_m(s_{k,m}^{(t-1)}, u_{k,m}^{(t)}), \quad
\bar{s}_{k,m}^{(t)} =
\hat{s}_{k,m}^{(t)} +
\mathrm{MLP}_m(\mathrm{LN}(\hat{s}_{k,m}^{(t)})).
\end{equation}
This update lets each block focus only on the semantic factor it represents,
making the representation more interpretable and less entangled.

To further encourage semantic separation, each block is projected onto a prototype memory shared across modalities:
\begin{equation}
s_{k,m}^{(t)} =
\mathrm{softmax}\!\left(
\frac{\bar{s}_{k,m}^{(t)}\mathbf{C}_m^\top}{\sqrt{d_c}}
\right)\mathbf{C}_m,
\end{equation}
where $\mathbf{C}_m \in \mathbb{R}^{K_m \times d_c}$ represents a compact dictionary
of $K_m$ learnable prototype vectors. Each prototype can be regarded as a basis
element for a specific semantic concept such as common colors or clothing
patterns. For instance, one prototype may represent ``red," another
``striped,” and others may represent garment shapes or texture types. During training, 
the prototype memories are optimized jointly with the network
via backpropagation, allowing each prototype to gradually converge to a
distinct semantic cluster that frequently appears in the dataset. The
softmax-weighted selection ensures that each block is discretely grounded to a
small subset of prototypes, rather than spreading over the entire feature
space. This effectively regularizes concept factor learning and prevents
semantic entanglement. Since both modalities share the same prototype memories
$\{\mathbf{C}_m\}_{m=1}^M$, the $m$-th block in images and text is guided to
align with a consistent set of semantic anchors. As a result, the refined slot
representations achieve not only part-level correspondence but also
concept-level consistency across modalities. This prototype projection thus
facilitates interpretable and fine-grained cross-modal matching (\textit{e.g.}, “red
shirt with stripes”).

After updating all blocks, the slot is reassembled:
$s_k^{(t)} = [s_{k,1}^{(t)}; \dots; s_{k,M}^{(t)}]$. A self-attention layer
is also applied among slots to reduce redundancy (\textit{e.g.}, two slots focusing on
the same part). After $T$ refinement iterations, we obtain:
\begin{equation}
\mathbf{S}_v^{(T)} = \{ s_k^{v,(T)} \}_{k=1}^K, \qquad
\mathbf{S}_t^{(T)} = \{ s_k^{t,(T)} \}_{k=1}^K.
\end{equation}
Each refined slot now corresponds to a meaningful body region and its text
description, while each block inside the slot encodes a distinct semantic
property. This enables accurate and interpretable matching of fine-grained
attributes between images and text.

\subsection{Training and Inference} \label{subsec:3.3}
The training objective aims to align visual and textual representations at different levels of granularity. At the global level, we optimize a bidirectional contrastive loss between the image embedding $g_v$ and the text embedding $g_t$ to capture overall cross-modal correspondence. This pulls paired samples closer in the embedding space while pushing non-matching pairs apart:
\begin{equation}
\mathcal{L}_{\text{global}}
= -\frac{1}{B}\sum_{i=1}^B 
\left[ 
\log \frac{\exp(\langle g_v^i, g_t^i \rangle / \tau)}
{\sum_{j=1}^B \exp(\langle g_v^i, g_t^j \rangle / \tau)}
+
\log \frac{\exp(\langle g_t^i, g_v^i \rangle / \tau)}
{\sum_{j=1}^B \exp(\langle g_t^i, g_v^j \rangle / \tau)}
\right],
\end{equation}
where $\tau$ is a learnable temperature and $B$ denotes the batch size.  

At the slot level, we align the refined slots across modalities, ensuring that the $k$-th slot in the visual branch corresponds to the same semantic region as the $k$-th slot in the textual branch:
\begin{equation}
\mathcal{L}_{\text{slot}}
= -\frac{1}{B}\sum_{i=1}^B \sum_{k=1}^K
\left[ 
\log \frac{\exp(\langle z_k^{v,i}, z_k^{t,i} \rangle / \tau_s)}
{\sum_{j=1}^B \exp(\langle z_k^{v,i}, z_k^{t,j} \rangle / \tau_s)}
\right].
\end{equation}

At the block level, we encourage alignment of disentangled concepts. Since each slot is further decomposed into concept blocks (\textit{e.g.}, color, texture, shape), we align corresponding blocks across modalities to ensure consistent matching of fine-grained cues:
\begin{equation}
\mathcal{L}_{\text{block}}
= -\frac{1}{B}\sum_{i=1}^B \sum_{k=1}^K \sum_{m=1}^M
\log \frac{\exp(\langle z_{k,m}^{v,i}, z_{k,m}^{t,i} \rangle / \tau_b)}
{\sum_{j=1}^B \exp(\langle z_{k,m}^{v,i}, z_{k,m}^{t,j} \rangle / \tau_b)}.
\end{equation}

To further enhance discriminability, we introduce identity supervision at both global and local levels. At the global level, the embeddings $g_v$ and $g_t$ are classified into identity labels, ensuring that they remain discriminative beyond cross-modal matching:
\begin{equation}
\mathcal{L}_{\text{ID}}^{\text{global}}
= -\frac{1}{B}\sum_{i=1}^B 
\big[ \log P(y_i \mid g_v^i) + \log P(y_i \mid g_t^i) \big].
\end{equation}
At the local level, each slot embedding ${z_k^u}$ ($u \in {v,t}$) is supervised with the same identity label, encouraging part-level features to preserve identity cues:
\begin{equation}
\mathcal{L}_{\text{ID}}^{\text{slot}}
= -\frac{1}{B}\sum_{i=1}^B \sum_{k=1}^K
\big[ \log P(y_i \mid z_k^{v,i}) + \log P(y_i \mid z_k^{t,i}) \big].
\end{equation}

Finally, we introduce a reconstruction objective to stabilize slot refinement. By reconstructing the original token features from aggregated slots, the model is encouraged to preserve the fine-grained information captured in the inputs:
\begin{equation}
\mathcal{L}_{\text{rec}}
= \frac{1}{B}\sum_{i=1}^B \big( \|\hat{\mathbf{X}}_v^i - \mathbf{X}_v^i\|_2^2
+ \|\hat{\mathbf{X}}_t^i - \mathbf{X}_t^i\|_2^2 \big).
\end{equation}

For clarity, we group these objectives into three components: alignment loss $\mathcal{L}_{\text{align}} = \mathcal{L}_{\text{global}} + \lambda_s \mathcal{L}_{\text{slot}} + \lambda_b \mathcal{L}_{\text{block}}$, identity loss $\mathcal{L}_{\text{ID}} = \mathcal{L}_{\text{ID}}^{\text{global}} + \mathcal{L}_{\text{ID}}^{\text{slot}}$, and reconstruction loss $\mathcal{L}_{\text{rec}}$. The overall training objective is then expressed as
\begin{equation}
\mathcal{L} = \mathcal{L}_{\text{align}} + \mathcal{L}_{\text{ID}} + \lambda_r \mathcal{L}_{\text{rec}}.
\end{equation}

During inference, a text query is matched with gallery images by combining similarities across multiple levels. Global embeddings provide identity-level cues, refined slots capture region-level correspondences, and block embeddings align fine-grained semantic attributes. The final retrieval score is computed as a weighted combination of these similarities, balancing overall identity with disentangled concept-level details.

\section{Experiments}

\subsection{Experiment Setting}

\subsubsection{Datasets} \label{subsubsec_exp:implement}
We conduct experiments on three widely used benchmarks for text-based person retrieval: CUHK-PEDES~\cite{CUHK-PEDES}, ICFG-PEDES~\cite{ICFG-PEDES}, and RSTPReid~\cite{RSTPReID}. CUHK-PEDES contains 40,206 images of 13,003 identities, each paired with two natural language descriptions, totaling 80,440 sentences. It is split into 34,054 images for training, 3,078 for validation, and 3,074 for testing. The dataset provides diverse sentence annotations that describe clothing, attributes, and contextual details. ICFG-PEDES is constructed from the ICFG-Person dataset and comprises 54,522 images of 4,102 identities, each annotated with one descriptive sentence. The official split includes 34,674 training images, 19,651 test images, and 197 for validation. Compared to CUHK-PEDES, the descriptions are shorter but still capture key appearance traits. RSTPReid is a large-scale benchmark focusing on more challenging scenarios. It consists of 20,505 images of 4,101 identities, with each image annotated by a single textual description. Following the official protocol, 8,000 identities are used for training, while the remaining are reserved for testing. The dataset features greater visual diversity, making retrieval particularly challenging.

\subsubsection{Implementation Details} \label{subsubsec_exp:implement}
Our model is built upon the CLIP~\cite{CLIP} ViT-L/14~\cite{dosovitskiy2020image} backbone with an image resolution of 384×128 pixels and a stride size of 16. The slot attention mechanism employs 8 slots with a slot dimension of 2,048, organized into 8 blocks of 256 dimensions each. We utilize a shared prototype memory with 512 prototypes across modalities to facilitate cross-modal part correspondence learning. The slot attention module performs 3 iterations during the iterative refinement process, and the temperature parameter for prototype assignment is set to 1.0. For training, we use the AdamW~\cite{kingma2014adam} optimizer with an initial learning rate of 3e-6. The overall loss is a weighted sum of all objectives, where $\lambda_s$ and $\lambda_b$ are set to 0.5 for the slot- and block-level losses, respectively, and $\lambda_r$ is set to 0.01. Data augmentation includes random horizontal flipping, padding with 10 pixels followed by random cropping, and random erasing with a scale range of [0.02, 0.4]. At inference, the final retrieval score is obtained from a weighted combination of global-, slot-, and block-level similarities, with all weights set to 1.

\begin{table*}[!t]
    \centering
    \setlength{\tabcolsep}{3pt} 
    \fontsize{15}{17}\selectfont
    \resizebox{1.0\textwidth}{!}{%
    \begin{tabular}{l|ccc|ccc|ccc}
    \toprule
    \textbf{Methods} & \multicolumn{3}{c|}{\textbf{CUHK-PEDES}} & \multicolumn{3}{c|}{\textbf{ICFG-PEDES}} & \multicolumn{3}{c}{\textbf{RSTPReid}} \\[-0.3ex]  
    \cmidrule(lr){2-4} \cmidrule(lr){5-7} \cmidrule(lr){8-10}
     & \multicolumn{1}{c}{R@1} & \multicolumn{1}{c}{R@5} & \multicolumn{1}{c|}{R@10} 
     & \multicolumn{1}{c}{R@1} & \multicolumn{1}{c}{R@5} & R@10 
     & \multicolumn{1}{c}{R@1} & \multicolumn{1}{c}{R@5} & \multicolumn{1}{c}{R@10} \\ 
    \midrule
    GNA-RNN~\cite{li2017person} & 19.05 & - & 53.64 & - & - & - & - & - & -\\
    CMPM/C~\cite{zhang2018deep} & 49.37 & 71.69 & 79.27 & 43.51 & 65.44 & 74.26 & - & - & - \\
    PMA~\cite{jing2020pose} & 53.81 & 73.54 & 81.23 & - & - & - & - & - & - \\
    TIMAM~\cite{sarafianos2019adversarial} & 54.51 & 77.56 & 84.78 & - & - & - & - & - & - \\
    ViTAA~\cite{wang2020vitaa} & 55.97 & 75.84 & 83.52 & 50.98 & 68.79 & 75.78 & - & - & - \\
    NAFS~\cite{gao2021contextual} & 59.94 & 79.86 & 86.70 & - & - & - & - & - & - \\
    SSAN~\cite{ding2021semantically} & 61.37 & 80.15 & 86.73 & 54.23 & 72.63 & 79.53 & 43.50 & 67.80 & 77.15 \\
    SRCF~\cite{suo2022simple} & 64.04 & 82.99 & 88.81 & 57.18 & 75.01 & 81.49 & - & - & -\\
    TIPCB~\cite{chen2022tipcb} & 64.26 & 83.19 & 89.10 & - & - & - & - & - & - \\
    SAF~\cite{li2022learning} & 64.13 & 82.62 & 88.40 & - & - & - & - & - & - \\
    IVT~\cite{shu2022see} & 65.59 & 83.11 & 89.21 & 56.04 & 73.60 & 80.22 & 46.70 & 70.00 & 78.80 \\
    CFine~\cite{yan2023clip} & 69.57 & 85.93 & 91.15 & 60.83 & 75.55 & 82.42 & 50.55 & 72.50 & 81.60 \\
    IRRA~\cite{jiang2023cross} & 73.38 & 89.93 & 93.71 & 63.46 & 80.24 & 85.82 & 60.20 & 81.30 & 88.20 \\ 
    PLOT~\cite{park2024plot} & 75.28 & 90.42 & 94.12 & 65.76 & 81.39 & 86.73 & 61.80 & 82.85 & 89.45 \\
    TBPS~\cite{cao2024empirical} & 73.54 & 88.19 & 92.35 & 65.05 & 80.34 & 85.47 & 62.10 & 81.90 & 87.75 \\
    CFAM~\cite{zuo2024ufinebench} & 75.60 & 90.53 & 94.36 & 65.38 & 81.17 & 86.35 & 62.45 & 83.55 & 91.10 \\
    MUM~\cite{zhao2024unifying} & 74.25 & 89.83 & 93.58 & 65.62 & 80.54 & 85.83 & 63.40 & 83.30 & 90.30 \\
    RDE~\cite{qin2024noisy} & 75.94 & 90.14 & 94.12 & 67.68 & 82.47 & 87.36 & 65.35 & 83.95 & 89.90 \\
    ICL~\cite{qin2025human} & 76.41 & 90.48 & 94.33 & 68.11 & 82.59 & 87.52 & 67.70 & 86.05 & 91.75 \\
    BAMG~\cite{cheng2024bamg} & 79.98 & 92.31 & 94.03 & 71.70 & 86.34 & 89.71 & 69.73 & 87.65 & 93.33 \\
    \midrule
    Ours & 77.21 & 91.85 & 95.63 & 67.81 & 83.29 & 87.62 & 67.84 & 85.72 & 91.98 \\ 
    \bottomrule
    \end{tabular}
        }
    \caption{Performance of text-to-image person re-identification methods on the three benchmark datasets.}
    \label{tab:sota}
\end{table*}

\subsection{Quantitative Results}
In Table~\ref{tab:sota}, we compare our DiCo with recent state-of-the-art methods on CUHK-PEDES~\cite{CUHK-PEDES}, ICFG-PEDES~\cite{ICFG-PEDES}, and RSTPReid~\cite{RSTPReID}. As shown in Table~\ref{tab:sota}, earlier global-matching methods (\textit{e.g.}, GNA-RNN~\cite{li2017person}, CMPM/C~\cite{zhang2018deep}) struggle to capture fine-grained details and thus exhibit limited retrieval performance. Recent methods leverage part-level alignment (\textit{e.g.}, TBPS~\cite{cao2024empirical}, CFAM~\cite{zuo2024ufinebench}, MUM~\cite{zhao2024unifying}, RDE~\cite{qin2024noisy}, ICL~\cite{qin2025human}, BAMG~\cite{cheng2024bamg}), demonstrating noticeable improvements in localization and discriminability. Recently, while BAMG~\cite{cheng2024bamg} improves performance by exploiting stronger part priors, their reliance on an additional segmentation model introduces extra training cost and restricts generalizability. In contrast, DiCo learns part decomposition and semantic concept representations jointly in an end-to-end manner, enabling more flexible and interpretable hierarchical alignment across global, body-slot, and concept-block levels. Thanks to this design, DiCo achieves competitive performance across all three benchmarks, with 77.21\% R@1 on CUHK-PEDES, 67.81\% on ICFG-PEDES, and 67.84\% on RSTPReid, clearly demonstrating the effectiveness of our modeling strategy in fine-grained text-to-image person retrieval. We note that ICL~\cite{qin2025human} can achieve slightly higher accuracy by leveraging test-time human-centered interaction and external MLLM knowledge to adaptively refine queries, especially when the original descriptions are ambiguous. In contrast, DiCo is a fully offline and interaction-free framework, and the small performance gap reflects a trade-off between interactive adaptability and scalable structured representation learning.

\begin{figure}[t!]
  \centering
  \includegraphics[width=0.9\linewidth]{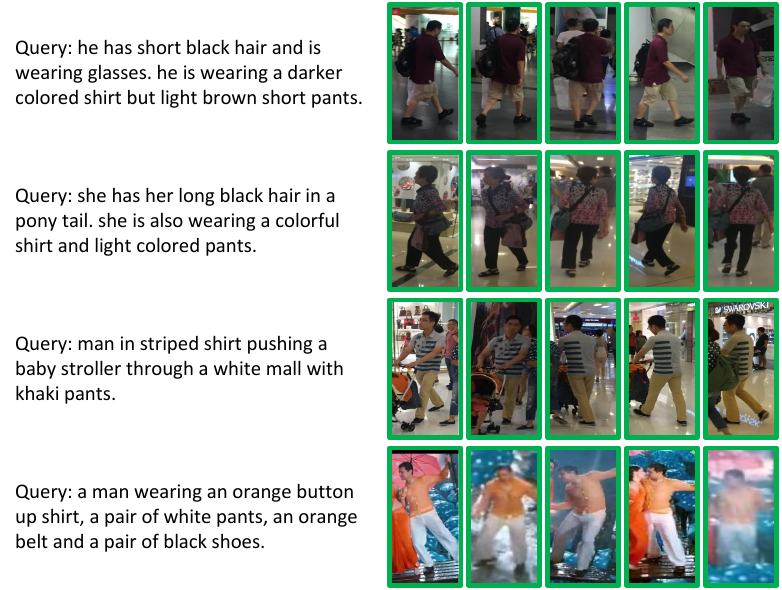}
  \caption{
  Qualitative results on CUHK-PEDES~\cite{CUHK-PEDES}. Each query description (left) is shown with the top-5 retrieved images (right). Green boxes indicate correct matches.
  }
  \label{fig:qualitative_results}
\end{figure}

\begin{figure}[t!]
  \centering
  \includegraphics[width=0.9\linewidth]{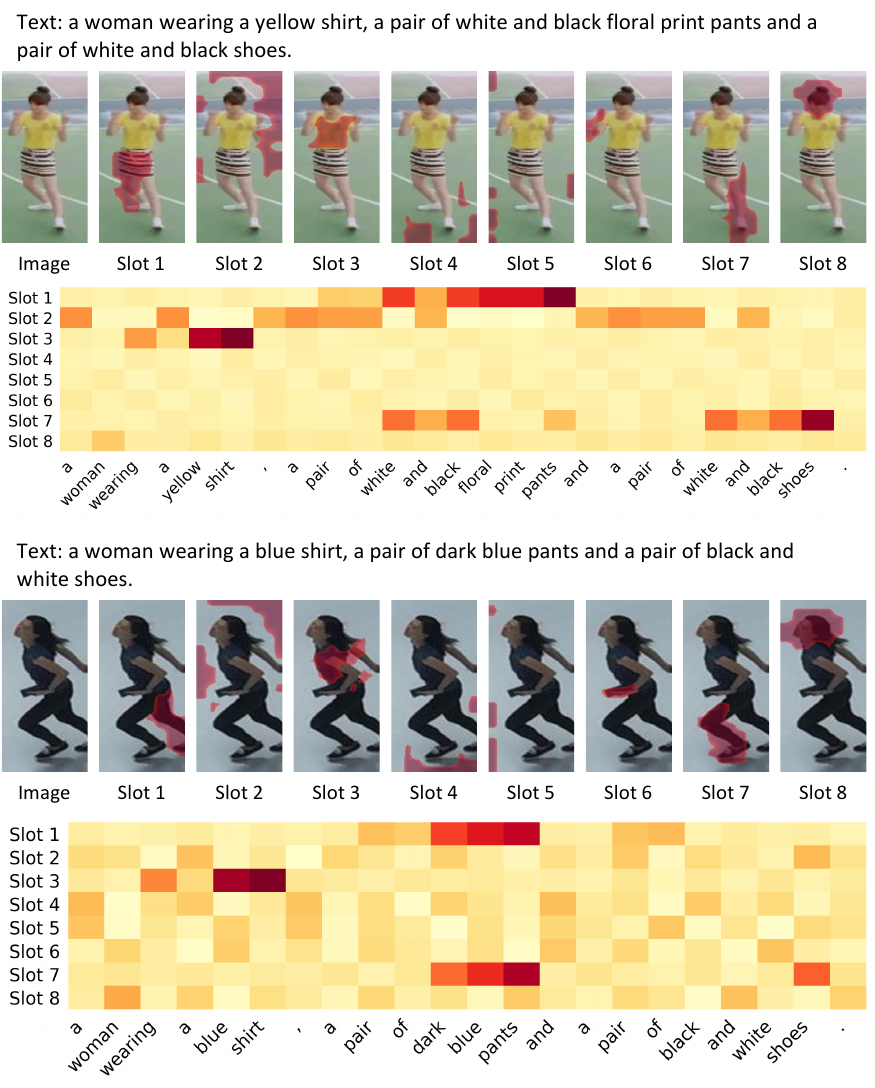}
  \caption{Visualization of slot-level attention for text queries and corresponding images. Each slot attends to distinct body regions and aligns with relevant words in the query, effectively capturing fine-grained attributes such as color, clothing type, and shoes.}
  \label{fig:slot_attention_vis}
\end{figure}

\begin{figure}[t!]
  \centering
  \includegraphics[width=1.0\linewidth]{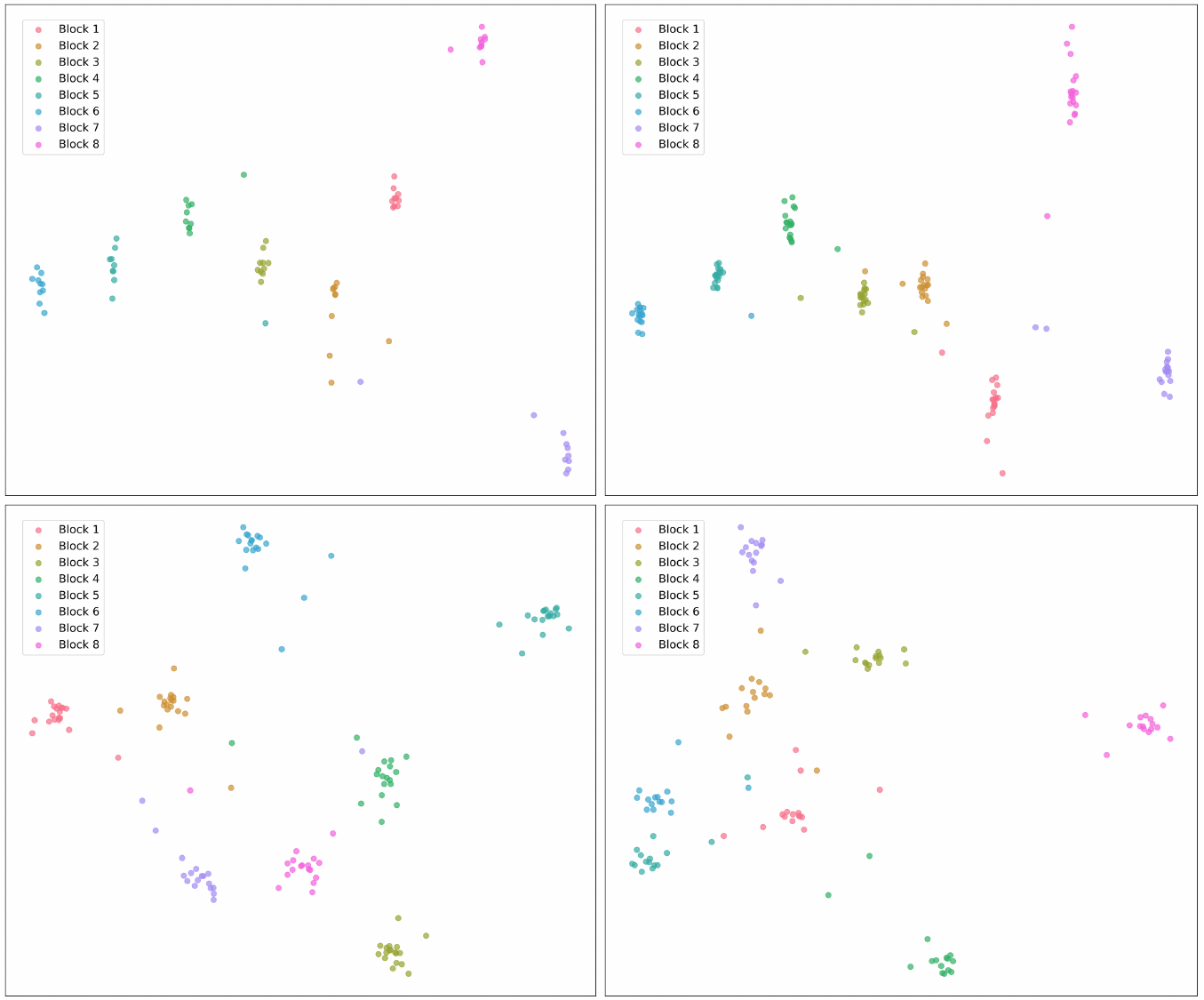}
  \caption{t-SNE~\cite{JMLR:v9:vandermaaten08a} visualization of concept block embeddings. 
        We project the learned concept representations into a 2D space using t-SNE to 
        examine whether different blocks capture distinct semantic subspaces. 
        Each scatter plot represents a different sampled subset of concept embeddings, 
        where points are colored by their corresponding block index. 
        Across multiple views, we observe consistently separated clusters among the 
        eight blocks, indicating that DiCo learns specialized and semantically diverse 
        concept representations rather than collapsing into overlapping embedding regions.}
\label{fig:block_tsne}
\end{figure}

\begin{figure}[t!]
  \centering
  \includegraphics[width=0.9\linewidth]{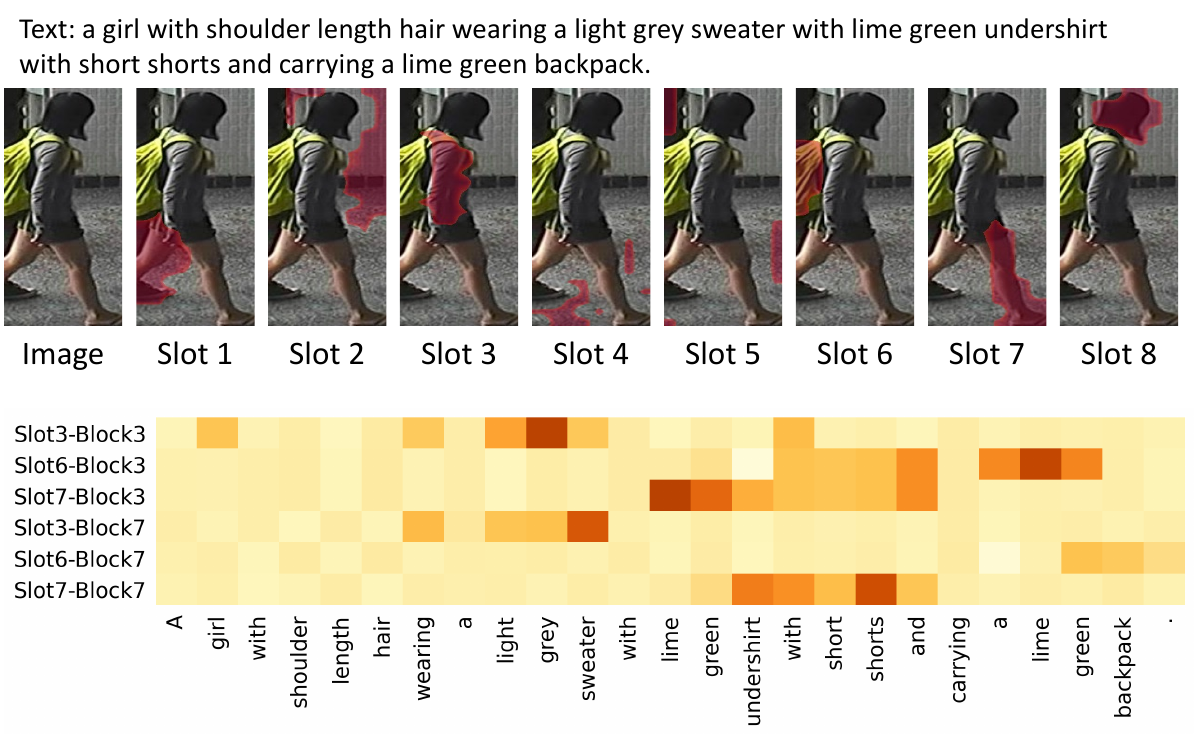}
  \caption{Qualitative visualization of slot- and block-level alignments on CUHK-PEDES~\cite{CUHK-PEDES} test set. Top: slot attention maps overlaid on the image given a text query, where each slot captures a distinct body region. Bottom: block-to-text interactions showing that Block~3 consistently attends to color-related words (\textit{e.g.}, \textit{lime}, \textit{green}), while Block~7 focuses on clothing-related words (\textit{e.g.}, \textit{sweater}, \textit{undershirt}). These results demonstrate that our disentangled block design captures semantically consistent concepts across modalities.}
  \label{fig:block_vis}
\end{figure}

\subsection{Qualitative Results}
\paragraph{Retrieval Results} We demonstrate that our model retrieves the correct person images from natural language queries, as shown in Fig.~\ref{fig:qualitative_results}. Given natural language queries, our model retrieves the correct person images within the top-5 results across diverse scenarios. As shown, the retrieved images not only match the coarse appearance described in the text but also capture fine-grained attributes such as specific clothing colors, patterns, and accessories. These examples highlight the model’s ability to disentangle semantic concepts and align subtle visual cues with textual descriptions, demonstrating its robustness in challenging cases where individuals share similar overall appearance.

\paragraph{Slot-Level Attention} In Fig.~\ref{fig:slot_attention_vis}, we provide visualizations of the learned slot-level attention to better understand how our model captures fine-grained correspondences between text and image. Each slot attends to distinct regions of the person image, such as the shirt, pants, or shoes, while simultaneously focusing on semantically related words in the textual description. The heatmaps depict the attention distribution between the $8$ slots (rows) and the variable-length text tokens (columns). For example, slots specialize in attributes like \textit{“yellow shirt”}, \textit{“floral pants”}, or \textit{“blue shirt”}, consistently linking localized visual cues with their linguistic counterparts. These results highlight that the proposed disentangled slot representations partition the image into semantically meaningful parts. They also capture fine-grained alignments with textual tokens, enabling precise cross-modal matching without the need for part-level annotations.

\paragraph{Block-Level Clustering}
We further validate the disentanglement ability of DiCo by analyzing block embeddings 
on CUHK-PEDES~\cite{CUHK-PEDES}, as shown in Fig.~\ref{fig:block_tsne}. 
Specifically, we randomly sample multiple sets of pedestrian images and visualize 
the embeddings of the eight blocks using t-SNE~\cite{JMLR:v9:vandermaaten08a}. 
Across different visualization instances, we observe consistently clear clustering 
patterns where features from different samples group together based on shared semantic 
concepts, even when the corresponding images depict distinct individuals. 
This indicates that each block captures a stable, concept-specific representation rather 
than encoding appearance variations tied to identity. 
Such consistent clustering behavior across multiple examples demonstrates the robustness 
of the proposed block-level design in learning disentangled and interpretable concepts, 
facilitating fine-grained cross-modal alignment.

\paragraph{Block-to-Text Alignment} The disentangled design of our framework enables different slots and blocks to capture semantically consistent factors, as shown in Fig.~\ref{fig:block_vis}. In the top section, slot-level attention highlights distinct body regions in the image given the text query, demonstrating the ability of slots to localize part-level semantics. In the bottom section, we visualize block-to-text interactions, where Block 3 consistently attends to color-related words (\textit{e.g.}, \textit{lime}, \textit{green}), while Block~7 aligns with clothing-related words (\textit{e.g.}, \textit{sweater}, \textit{undershirt}), regardless of the slot they belong to. These results confirm that our block-wise disentanglement encourages specialization into meaningful concepts, which in turn facilitates precise and interpretable cross-modal alignment.

\begin{table}[t]
  \centering
  \setlength{\tabcolsep}{3pt} 
  \renewcommand{\arraystretch}{1.3}
  \resizebox{0.98\textwidth}{!}{%
  \begin{tabular}{c|l|ccccc|ccc}
  \toprule
  \multirow{2}{*}{No.} & \multirow{2}{*}{Variant} 
  & \(\mathcal{L}_{\text{global}}\) & \(\mathcal{L}_{\text{slot}}\) & \(\mathcal{L}_{\text{block}}\) 
  & \(\mathcal{L}_{\text{ID}}^{\text{local}}\) & \(\mathcal{L}_{\text{rec}}\)
  & \multicolumn{3}{c}{CUHK-PEDES} \\
  \cline{8-10}
   &  &  &  &  &  &  
  & R@1 & R@5 & R@10 \\
  \midrule
  0 & Baseline (global alignment only)
    & \checkmark &  &  &  &  
    & 69.26 & 85.94 & 89.60 \\
  1 & + Slot-level alignment
    & \checkmark & \checkmark &  &  &  
    & 73.89 & 89.23 & 92.99 \\
  2 & + Block-level alignment
    & \checkmark & \checkmark & \checkmark &  &  
    & 75.25 & 90.73 & 94.58 \\
  3 & + Slot ID supervision
    & \checkmark & \checkmark & \checkmark & \checkmark &  
    & \underline{76.89} & \underline{91.48} & \underline{95.25} \\
  4 & + Reconstruction regularization
    & \checkmark & \checkmark & \checkmark & \checkmark & \checkmark
    & \textbf{77.21} & \textbf{91.85} & \textbf{95.63} \\
  \bottomrule
  \end{tabular}%
  }
  \caption{Ablation on CUHK-PEDES~\cite{CUHK-PEDES} using our loss components. We incrementally add slot- and block-level alignment, local (slot) identity supervision, and reconstruction. \textbf{Bold} denotes the best and \underline{underline} denotes the second best.}
  \label{tab:ablation_cuhk}
\end{table}

\begin{table}[t]
  \centering
  \setlength{\tabcolsep}{5pt}
  \renewcommand{\arraystretch}{1.2}
  \begin{tabular}{c|c|c|ccc}
  \toprule
  \#Slots & \#Blocks & \#Prototypes & R@1 & R@5 & R@10 \\
  \midrule
   4 & 8  & 256 & 76.01 & 89.45 & 94.77 \\
   8 & 8  & 256 & \textbf{77.21} & \textbf{91.85} & \textbf{95.63} \\
  12 & 8  & 256 & \underline{76.98} & \underline{91.81} & \underline{95.22} \\
  \midrule
   8 & 4  & 256 & 77.01 & 91.33 & 95.11 \\
   8 & 8  & 256 & \textbf{77.21} & \textbf{91.85} & \textbf{95.63} \\
   8 & 12 & 256 & \underline{77.10} & \underline{91.56} & \underline{95.62} \\
  \midrule
   8 & 8  & 64  & 76.62 & 90.89 & 94.91 \\
   8 & 8  & 128 & \underline{77.01} & \underline{91.24} & \underline{95.55} \\
   8 & 8  & 256 & \textbf{77.21} & \textbf{91.85} & \textbf{95.63} \\
  \bottomrule
  \end{tabular}
  \caption{Ablation study on the CUHK-PEDES~\cite{CUHK-PEDES} dataset with varying numbers of slots, blocks, and prototypes. We vary one factor while fixing the others (default: 8 slots, 8 blocks, 256 prototypes). \textbf{Bold} denotes the best and \underline{underline} denotes the second best.}
  \label{tab:ablation_slots_blocks_prototypes}
\end{table}

\subsection{Ablation Study}

\paragraph{Module effectiveness} 
We conduct an ablation study on CUHK-PEDES~\cite{CUHK-PEDES} to analyze the contribution of each component in our framework, as summarized in Table~\ref{tab:ablation_cuhk}. 
The baseline model, which adopts the CLIP ViT-L backbone with only global alignment ($\mathcal{L}_{\text{global}}$), achieves 69.26\% R@1.
Introducing slot-level alignment ($\mathcal{L}_{\text{slot}}$) brings a clear improvement (+4.63\% R@1), verifying the effectiveness of explicitly modeling part-to-text correspondences. 
Adding block-level alignment ($\mathcal{L}_{\text{block}}$) further enhances fine-grained semantic reasoning, improving R@1 to 75.25\%. 
With additional slot identity supervision ($\mathcal{L}_{\text{ID}}^{\text{local}}$), the model becomes more discriminative, achieving 76.89\% R@1. 
Finally, applying reconstruction regularization ($\mathcal{L}_{\text{rec}}$) stabilizes learning and encourages consistent concept representations, reaching the best performance of 77.21\% R@1, 91.85\% R@5, and 95.63\% R@10. 
These results consistently demonstrate that (1) slot- and block-level alignments enhance multi-granular semantic grounding, (2) local identity supervision strengthens pedestrian distinctiveness, and (3) reconstruction improves robustness, collaboratively yielding substantial improvements over the strong CLIP-based baseline.

\paragraph{Number of slots, blocks, and prototypes} We analyze the effect of varying the number of slots, blocks, and prototypes on CUHK-PEDES, as shown in Table~\ref{tab:ablation_slots_blocks_prototypes}. Increasing the number of slots from 4 to 8 yields a clear gain, as additional slots allow the model to capture body regions more precisely. However, further increasing to 12 leads to slight performance drops, suggesting that too many slots may introduce redundancy and reduce discriminability. A similar pattern is observed with blocks: using only 4 blocks limits the ability to represent diverse concepts, while 12 blocks dilute the representation and slightly degrade performance. The best balance is achieved with 8 blocks, which provide sufficient granularity without unnecessary complexity. For prototypes, enlarging the dictionary from 64 to 256 steadily enhances performance by providing richer semantic anchors. These results highlight that an appropriate balance of slots, blocks, and prototypes is essential, with 8 slots, 8 blocks, and 256 prototypes giving the most effective configuration.

\begin{table}[t]
  \centering
  \setlength{\tabcolsep}{5pt}
  \renewcommand{\arraystretch}{1.2}
  \begin{tabular}{c|c|c|ccc}
  \toprule
  $\lambda_s$ & $\lambda_b$ & $\lambda_r$ & R@1 & R@5 & R@10 \\
  \midrule
  0.0 & 0.5 & 0.1 & 73.88 & 88.51 & 91.01 \\
  0.5 & 0.0 & 0.1 & 75.55 & 90.75 & 93.23 \\
  0.5 & 0.5 & 0.0 & 76.68 & 91.21 & 94.89 \\
  0.5 & 0.5 & 0.1 & \textbf{77.21} & \textbf{91.85} & \textbf{95.63} \\
  1.0 & 0.5 & 0.1 & \underline{77.10} & \underline{91.75} & \underline{95.52} \\
  0.5 & 1.0 & 0.1 & 76.51 & 90.77 & 94.49 \\
  0.5 & 0.5 & 0.2 & 76.20 & 90.92 & 94.60 \\
  \bottomrule
  \end{tabular}
  \caption{Ablation study on CUHK-PEDES~\cite{CUHK-PEDES} with different values of $\lambda_s$, $\lambda_b$, and $\lambda_r$. \textbf{Bold} indicates the best result and \underline{underline} indicates the second best.}
  \label{tab:ablation_loss_weights}
\end{table}

\paragraph{Loss weight} We analyze the effect of varying the loss weights $\lambda_s$, $\lambda_b$, and $\lambda_r$ in Table~\ref{tab:ablation_loss_weights}. Removing either the slot-level or block-level loss ($\lambda_s=0$ or $\lambda_b=0$) leads to a clear performance drop, confirming the importance of both part-level and concept-level alignment. Similarly, discarding the reconstruction loss ($\lambda_r=0$) slightly reduces retrieval accuracy, showing its role in preserving fine-grained details during slot refinement. The best trade-off is achieved with $\lambda_s=0.5$, $\lambda_b=0.5$, and $\lambda_r=0.1$, which balance alignment and regularization, yielding the highest performance across all metrics.

\begin{figure}[t]
  \centering
  \includegraphics[width=1.0\linewidth]{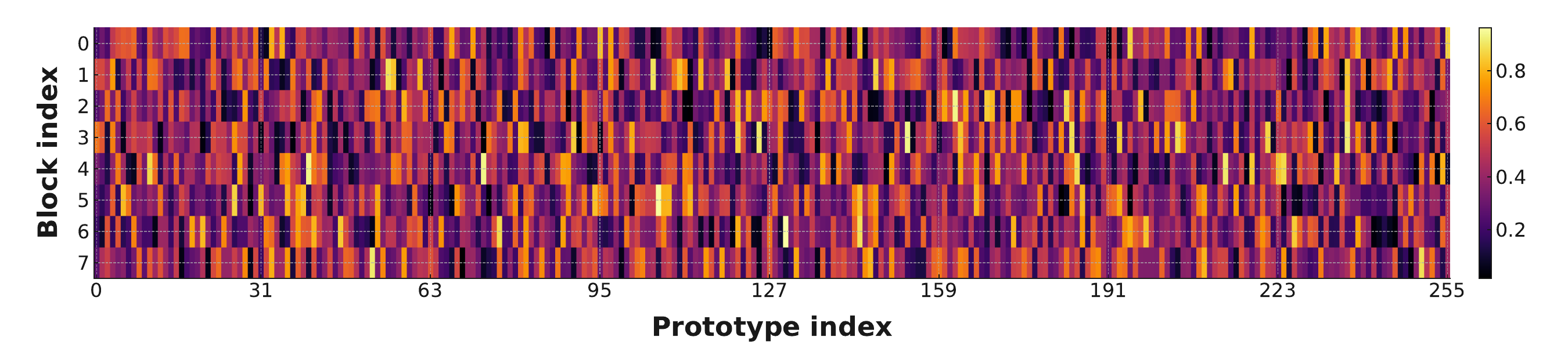}
  \caption{Prototype coverage on CUHK-PEDES. We visualize the average activation of each prototype 
across the test set. Each row corresponds to a block index and each column indicates a prototype 
within the corresponding concept memory. The wide spread of active prototypes across blocks 
demonstrates that our learned concept memories are effectively utilized to represent diverse 
fine-grained semantics.}
  \label{fig:proto_coverage}
\end{figure}

\paragraph{Prototype memory coverage} 
To further validate the effectiveness of our concept memory design, in Fig.~\ref{fig:proto_coverage}, we analyze the usage patterns of 
prototypes across the entire CUHK-PEDES test set. Each row corresponds to a block index ($m = 1,\dots,8$) and each column 
represents a prototype index ($k = 1,\dots,256$) within the corresponding concept memory $C_m$. Each 
cell indicates the average activation weight of a prototype when retrieving pedestrian images. 
We observe that a wide range of prototypes exhibit non-negligible activation responses rather than 
collapsing to only a few dominant slots, showing that different prototypes are meaningfully involved 
in representing diverse semantic concepts. Moreover, the distribution is well spread across all 
eight blocks, suggesting that the proposed hierarchical alignment encourages semantically 
complementary behaviors among concept memories $C_m$. These results confirm that the learned 
prototypes serve as distinct and diverse semantic cues, aligning with our goal of improving both 
interpretability and fine-grained reasoning in text-to-image person retrieval.

\begin{figure}[t]
  \centering
  \includegraphics[width=1.0\linewidth]{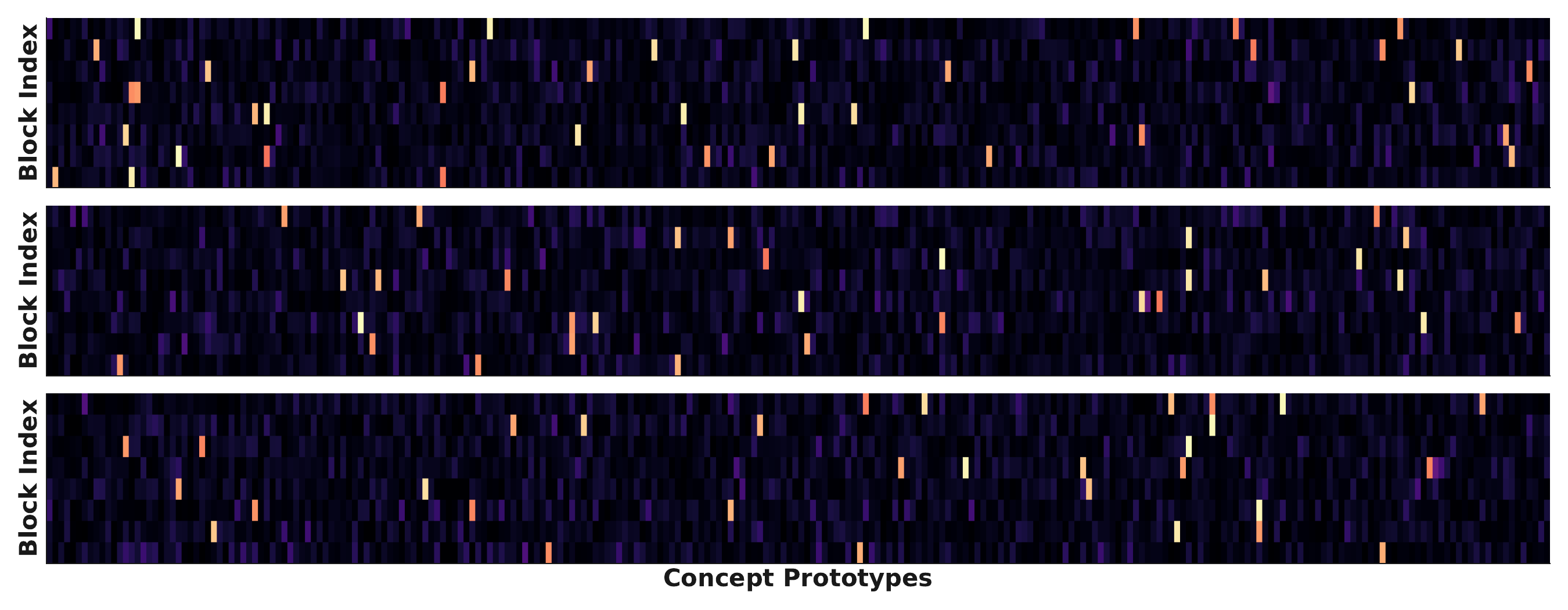}
  \caption{Prototype attention visualization. 
We show attention responses from three sampled blocks to all prototypes in the 
concept memories. Each row denotes a block index and each column a prototype. 
The activations indicate that blocks attend to multiple prototypes rather than 
performing a single hard selection, confirming the flexibility and semantic 
expressiveness of the learned concept memory.}
  \label{fig:proto_attention}
\end{figure}

\paragraph{Prototype attention behavior}
To investigate how DiCo interacts with the concept memory during retrieval, 
we visualize the attention weights from several blocks to the prototype vectors, 
as shown in Fig.~\ref{fig:proto_attention}. Each row corresponds to a block index 
and each column represents a prototype from the corresponding concept memory. 
We observe that the attention distributions are not one-hot but instead involve 
a soft combination of multiple prototypes. This demonstrates that the blocks do 
not rely on a single prototype, but flexibly integrate multiple semantic cues 
depending on the textual description and pedestrian appearance. The diversity 
of activated prototypes across different blocks confirms that DiCo leverages 
semantically complementary concepts, enabling richer and more interpretable 
fine-grained alignment.

\section{Limitation and Future Work}
While our framework has shown strong performance and interpretability, several aspects suggest promising directions for further research. For instance, the slot representations, though generally effective in capturing person-related regions, sometimes also respond to background areas. Incorporating guiding mechanisms or controllable constraints may encourage more precise localization of foreground semantics. In addition, although the block-wise design successfully captures concept-level factors such as color or clothing type, achieving fully disentangled and universally interpretable blocks remains an open challenge. Exploring stronger inductive biases or structured supervision could further enhance the semantic clarity of each block, leading to even more explainable and controllable cross-modal alignment.

\section{Conclusion}
In this work, we have proposed Disentangled Concept Representation (DiCo), a novel framework for text-to-image person re-identification that has addressed the challenges of modality gap and fine-grained correspondence. By decomposing representations into shared part-level slots and disentangled concept blocks, our method has enabled hierarchical alignment across global, part, and concept levels without requiring explicit part annotations. Extensive experiments on three public benchmarks have demonstrated that DiCo has achieved competitive performance with state-of-the-art methods, while providing interpretable and fine-grained cross-modal matching.

\section{Acknowledgements}

% This research was supported by Culture, Sports and Tourism R\&D Program through the Korea Creative Content Agency grant funded by Ministry of Culture, Sports and  Tourism in 2024 (Project Name : Developing Professionals for R\&D in Contents Production Based on Generative Ai and Cloud, Project Number : RS-2024-00352578, Contribution Rate: 00%)

This work was supported by the National Research Foundation of Korea (NRF) grant funded by the Korea government (MSIT) (RS-2024-00355008). It was also supported by the Culture, Sports and Tourism R\&D Program through the Korea Creative Content Agency grant funded by the Ministry of Culture, Sports and Tourism in 2024 (Project Name: Developing Professionals for R\&D in Contents Production Based on Generative AI and Cloud, Project Number: RS-2024-00352578, Contribution Rate: 30\%). In addition, this research was supported by the Chung-Ang University Research Scholarship Grants in 2024.

\bibliographystyle{elsarticle-num} 
\bibliography{ref} 

\end{document}